\PassOptionsToPackage{pdftex,dvipsnames}{xcolor}
\documentclass[letterpaper, 10 pt, conference]{ieeeconf}  

\IEEEoverridecommandlockouts                             
\usepackage{amsmath}
\usepackage{amssymb}
\let\labelindent\relax
\usepackage{enumitem}

\usepackage{amsmath,amsfonts}
\usepackage[font=footnotesize]{caption}
\usepackage{subcaption}
\usepackage{algorithm}
\usepackage{algpseudocode}
\usepackage{array}
\usepackage{textcomp}
\usepackage{url}
\usepackage{graphicx}
\usepackage{tabularray}
\usepackage{float}
\usepackage{bm}
\usepackage{mathtools}
\usepackage{stfloats}
\usepackage{mathrsfs}
\usepackage{multirow}
\usepackage{makecell}
\usepackage{vcell}
\usepackage{booktabs}
\usepackage{diagbox}
\usepackage{csquotes}
\usepackage{tabularx}
\usepackage{tikz}
\usepackage{lipsum}
\usepackage{url}
\usepackage{schemata}
\usepackage{verbatim}
\usepackage{pifont}

\usepackage{cite}
\makeatletter
\let\NAT@parse\undefined
\makeatother
\usepackage{hyperref}
\hypersetup{
colorlinks=true,
linkcolor=blue,
filecolor=magenta,
urlcolor=blue,
}
\hypersetup{breaklinks=true}
\title{\LARGE \bf
Containing Behavioral Cascades from Manipulated Claims in LLM-Powered Multi-Robot Systems}

\author{Waleed Bin Khalid and Byung-Cheol Min
\thanks{Waleed Bin Khalid and Byung-Cheol Min are with the
SMART Lab, Department of Computer Science, Luddy School of
Informatics, Computing, and Engineering, Indiana University
Bloomington, Bloomington, IN, USA.
Corresponding authors: Waleed Bin Khalid and Byung-Cheol Min
(e-mail: \{wakhalid,minb\}@iu.edu).}%
}
\begin{document}

\maketitle
\thispagestyle{empty}
\pagestyle{empty}


\begin{abstract}
Large language model (LLM)-powered multi-robot systems are vulnerable to
semantic manipulation: an accepted false world-state claim can trigger a
fleet-wide behavioral cascade, causing unnecessary replanning, increased path
costs, congestion, or apparent mission infeasibility. Conditioning on a
successful manipulation, we propose an active verification framework that
contains its downstream effects before they propagate across the fleet. A
dedicated verification module generates a structured
\textit{Verify--Adapt--Hold} plan: selected robots inspect consequential
regions, a limited subset provisionally adapts when necessary, and the
remaining robots retain their trusted plans. We evaluate the framework in a
multi-robot transportation environment using injected false obstacle claims
across different impacts and team sizes. Evaluation measures cascade
containment, Sum-of-Costs, makespan, and coverage ratio.
Results show that treating post-compromise verification as a team-level
planning problem, rather than a binary trust decision, effectively limits the
cascading physical consequences of semantic manipulation. Additional materials are available at \url{https://sites.google.com/view/vahframework}.
\end{abstract}

\section{Introduction}
Large language models (LLMs) increasingly serve as high-level robotic planners
that translate natural-language instructions and world-state descriptions into
structured task and coordination decisions~\cite{driess2023palme,ahn2022saycan,
liang2023code,wake2023gpt4robot}. In multi-robot systems, LLM-based planners can
decompose tasks, form robot coalitions, allocate tasks across the fleet, and
generate structured multi-robot plans~\cite{kannan2023smart}.

However, malicious inputs or compromised external observations can manipulate
the reasoning and actions of LLM-based systems
~\cite{perez2022ignore,greshake2023not,yang2024watch}.
When such a system serves as a robot planner, the resulting decisions can alter
the behavior of one or more robots. In a shared reservation-based planner,
the effects may then propagate beyond robots whose trajectories intersect the
claimed region: rerouting one robot can alter the feasible paths available to
robots planned after it. We call this fleet-wide propagation of claim-induced
behavior a \emph{behavioral cascade}. Such cascades can cause unnecessary
replanning, increased path costs, and mission infeasibility.

\begin{figure}[t]
  \centering
  \includegraphics[width=0.9\columnwidth]{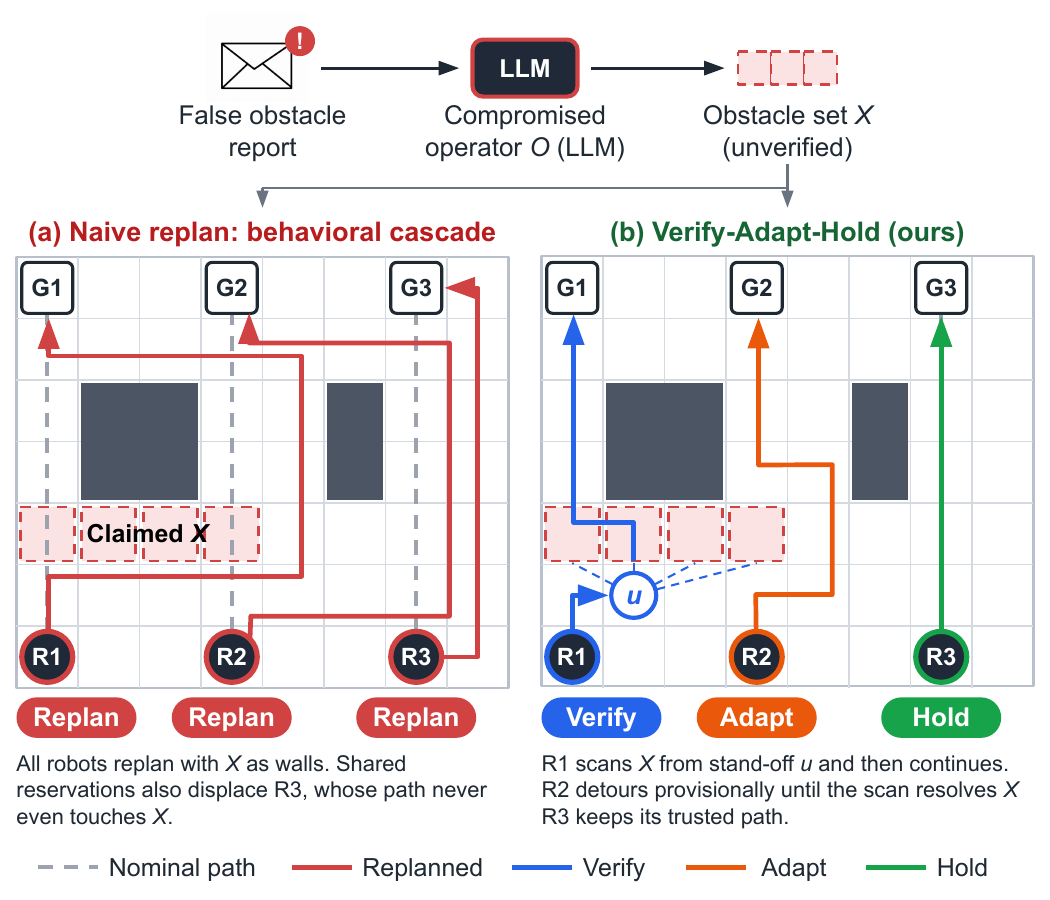}
  \vspace{-5pt}
  \caption{\scriptsize
  (a) Under Naive Replan, all robots treat the false obstacle claim
  \(\mathcal{X}\) as blocked and replan, propagating route changes even to robots whose paths do not
  intersect \(\mathcal{X}\).
  (b) Under Verify--Adapt--Hold, a selected robot verifies the claim from a
  stand-off, threatened robots provisionally adapt around unresolved claimed
  cells, and unaffected robots retain their trusted trajectories until physical
  evidence resolves the claim.}
  \vspace{-15pt}
  \label{fig:teaser}
\end{figure}

Semantic manipulation can arise through direct or indirect prompt injection
~\cite{perez2022ignore,greshake2023not}, malicious triggers introduced through
external observations~\cite{yang2024watch}, adversarial perturbations to language
inputs~\cite{zhu2023promptbench}, cyber attacks that corrupt or disrupt communication in multi-robot systems~\cite{lee2021distributed,yeke2025raven}, or operator
error. Despite their different origins, these mechanisms can introduce incorrect
information into downstream decision making. In a coupled multi-robot planning
system, the resulting effects propagate beyond the directly affected robot.

Existing language-model security research has characterized manipulation
and jailbreak vulnerabilities~\cite{wei2023jailbroken} and proposed or
benchmarked defenses against them~\cite{robey2023smoothllm,armstrong2023prompt}.
However, the downstream response after a semantic manipulation succeeds remains underexplored. In particular,
it remains unclear how a robot team should acquire physical evidence while
continuing its mission and limiting unnecessary behavioral propagation.

Figure~\ref{fig:teaser} illustrates the behavioral cascade and our proposed
response. Instead of either accepting and replanning or ignoring an untrusted claim, we formulate the response as a team-level planning
problem. Our \emph{Verify--Adapt--Hold} framework treats the claim as
provisional: selected robots verify consequential regions, robots whose trusted
trajectories conflict with unresolved claims provisionally adapt, and
unaffected robots continue their existing plans. This paper conditions on a \emph{successful manipulation} of the high-level LLM
planner and focuses on its consequences. The main contributions of this paper are:

\begin{itemize}

\item We formulate post-compromise handling of manipulated world-state claims
in LLM-powered multi-robot systems as a team-level planning problem and
introduce a \emph{Verify--Adapt--Hold} framework that jointly coordinates
verification and mission execution.

\item We develop a decision-relevance-aware verification strategy that uses
LLM reasoning to select verification actions based jointly on the potential
mission impact of claimed regions and current robot trajectories. This
prioritizes consequential claims while minimizing disruption to ongoing
mission execution.

\item We extensively evaluate the framework across multiple team sizes and
claim-impact scenarios against immediate-replanning and verification-first
baselines. Using cascade containment, mission cost, makespan, and coverage, the
results show that the framework contains behavioral cascades while preserving
mission progress and verification coverage.

\end{itemize}

\section{RELATED WORK}

\textbf{Foundation Models in Robot Planning:}
Foundation models serve as direct policy generators or high-level reasoners
for robotic execution~\cite{brohan2023rt2,driess2023palme,ahn2022saycan},
translating natural-language instructions into structured plans or executable
programs~\cite{liang2023code,song2023llmplanner,rana2023sayplan}. In
multi-robot systems, LLM-based planning has been explored under centralized,
decentralized, and collaborative coordination architectures
~\cite{chen2024scalable,mandi2024roco,zhang2023building,kannan2023smart}.
Large-scale warehouse systems similarly require fleet-level task, routing,
and coordination mechanisms~\cite{wurman2008coordinating,
fragapane2021planning}. Our framework adopts a centralized high-level
operator for task allocation and \(A^*\) for trajectory execution, augmented
by a verification module at the planning level that leaves allocation logic
unchanged.

\textbf{Semantic Manipulation:}
Foundation-model planners are vulnerable to adversarial inputs that redirect
their reasoning~\cite{perez2022ignore,greshake2023not,zhu2023promptbench},
including direct and indirect prompt injection through malicious instructions
or externally supplied content~\cite{perez2022ignore,greshake2023not}.
LLM-based agents are also vulnerable to malicious triggers appearing in user
queries or intermediate observations returned by the environment
~\cite{yang2024watch}. Existing work has characterized jailbreak vulnerabilities
~\cite{wei2023jailbroken} and proposed or benchmarked defenses such as randomized
input perturbation and prompt-injection countermeasures
~\cite{robey2023smoothllm,armstrong2023prompt}. These defenses reduce, but do not
eliminate, successful manipulation. Our work complements them by containing the
physical consequences after a false claim has been accepted.

\textbf{Authority-Framed Manipulation:}
Recent work shows that LLMs are susceptible to persuasive framing, including
appeals to authority and expert endorsement, which can increase compliance with
otherwise restricted requests~\cite{zeng2024persuade,meincke2026persuading}.
Authority cues can therefore constitute a manipulation surface when an
unverified claim is presented as originating from a credible expert or official.
Related work has further demonstrated authority-citation-driven jailbreak
attacks that exploit LLMs' tendency to assign greater trust to apparently
authoritative information~\cite{yang2024darkside}. In this work, we instantiate this threat by presenting a false obstacle report as originating
from an apparent safety authority. We condition on the operator accepting the
claim and study containment of its downstream physical effects.

\textbf{Behavioral Cascades in Multi-Robot Systems:}
Multi-robot path planning couples robot trajectories through collision-avoidance
constraints, so the feasible path of one robot can depend on the paths assigned
to others~\cite{stern2019multi,li2021lifelong}. In our shared reservation-based
planner, this coupling creates a mechanism through which replanning one robot
can constrain robots planned afterward, including robots whose trajectories do
not directly intersect the perturbed region. LLM-based multi-robot planning has
also demonstrated the ability to coordinate decisions across multiple robots
~\cite{chen2024scalable,mandi2024roco}. However, the downstream propagation of
behavior after a high-level operator accepts a manipulated world-state claim
remains underexplored, and our work targets this gap.

\textbf{Verification, Trust, and Active Perception:}
Runtime-assurance frameworks can monitor safety conditions during execution
and switch from high-performance components to verified-safe behavior when
necessary~\cite{desai2019soter}. Related formal-safety work has studied
formally verified exploration~\cite{anderson2021neurosymbolic} and formal
specification of learned components~\cite{seshia2018formal}. Active perception
directs sensing and motion to reduce uncertainty about the environment
~\cite{bajcsy2018revisiting}. Our work combines these ideas
by directing robots toward unverified claim regions, generating structured
assignments that preserve mission progress, and re-invoking the module only
when new physical evidence arrives.



\section{Problem Formulation}
\label{sec:formulation}

The mission is a multi-robot warehouse transportation task in which each robot
collects an assigned resource from a pickup location and delivers it to a
designated drop-off. The mission only succeeds when all robots deliver their resources.
 
\subsection{Environment and Fleet}

The warehouse is a finite two-dimensional grid of cells with width \(W\) and height \(H\), represented by \(\mathcal{G} = \{0,\dots,W-1\} \times \{0,\dots,H-1\}\). A ground-truth occupancy map labels each cell \(c\) as either free floor or physically occupied by a boundary wall, shelf, or pillar:
\vspace{-5pt}
\begin{equation}
\vspace{-5pt}
G_{\mathrm{true}}(c)
=
\begin{cases}
0, & \text{if } c \text{ is free},\\
1, & \text{if } c \text{ is physically occupied}.
\end{cases}
\end{equation}

The set of free cells is \(\mathcal{F} = \{c\in\mathcal{G}:G_{\mathrm{true}}(c)=0\}\). Pickup and drop-off cells are represented by \(\mathcal{P}\) and \(\mathcal{D}\), respectively, where \(\mathcal{P},\mathcal{D}\subseteq\mathcal{F}\). Multiple robots share a pickup location, but each robot has its own unique drop-off location. Both pickup and drop-off cells are traversable free cells. Time advances in discrete steps indexed by \(t\).

The fleet consists of \(N\) homogeneous robots, \(\mathcal{R}=\{R_0,\dots,R_{N-1}\}\), with positions \(p_i(t)\in\mathcal{F}\), where \(p_i(t)\neq p_j(t)\) for all \(i\neq j\). Motion is 4-connected, with waiting allowed and no diagonal motion. Each robot can carry one resource during its journey. A robot \(R_i\) can observe a cell \(c\) only if \(\left\|c-p_i(t)\right\|_{\infty}\leq\rho\), where \(\rho\) is the Chebyshev sensing radius, and the line of sight from \(p_i(t)\) to \(c\) is unobstructed. For robot $R_i$, let $s_i\in\mathcal{P}$ denote its assigned pickup location and $g_i\in\mathcal{D}$ its assigned drop-off location. Hence, its nominal mission at $t=0$ is $p_i(0)\rightarrow s_i\rightarrow g_i$. 



\subsection{Shared Space-Time Planner}
\label{sub:planner}

Let
\(\gamma_i=((p_i(t_0),t_0),\ldots,(p_i(t_f),t_f))\)
denote robot \(R_i\)'s planned space-time trajectory from time \(t_0\) to
\(t_f\). The fleet uses one shared reservation table
\(\mathcal{Z}=(\mathcal{Z}^{V},\mathcal{Z}^{E})\), which records occupied
cell-time pairs and transitions:
\vspace{-5pt}
\begin{equation}
\vspace{-5pt}
\begin{aligned}
\mathcal{Z}^{V}
&=
\left\{
(p_j(t),t)
\mid \gamma_j \text{ has been reserved}
\right\},\\
\mathcal{Z}^{E}
&=
\left\{
(p_j(t),p_j(t+1),t)
\mid \gamma_j \text{ has been reserved}
\right\}.
\end{aligned}
\end{equation}

These sets include all states and consecutive transitions of the reserved
trajectories. Robots are planned sequentially, and
\(\mathcal{Z}_{<i}\) denotes the contents of the shared table immediately
before \(R_i\) is planned, where \(<i\) refers to planning order rather than
robot index. The table is initially empty for the nominal plan, and each
generated trajectory is reserved before planning the next robot. During
replanning, trajectories that remain fixed are reserved first.

Let \(a_i\) denote robot \(R_i\)'s assignment, including its pickup \(s_i\),
drop-off \(g_i\), current role, and stand-off when applicable. Let
\(\mathcal{F}_i^{\mathrm{plan}}\) denote the cells currently treated as
traversable. Let STP denote the shared space-time planner. Starting at \(p_i(t_0)\), the planner returns, when feasible,
\vspace{-5pt}
\begin{equation}
\vspace{-5pt}
\gamma_i
=
\operatorname{STP}
\left(
(p_i(t_0),t_0),
a_i,
\mathcal{F}_i^{\mathrm{plan}},
\mathcal{Z}_{<i}
\right).
\end{equation}

The returned trajectory satisfies the remaining assignment and existing reservations. Its states and transitions satisfy \(p_i(t)\in\mathcal{F}_i^{\mathrm{plan}}\) and \(\left\|p_i(t+1)-p_i(t)\right\|_1\leq 1\), allowing 4-neighbor movement or waiting. Conflicts with reserved trajectories are prevented by requiring \((p_i(t),t)\notin\mathcal{Z}_{<i}^{V}\) and \((p_i(t+1),p_i(t),t)\notin\mathcal{Z}_{<i}^{E}\). The planner determines
only trajectory geometry and timing; it does not modify \(a_i\) and reports
failure if no feasible trajectory is found.

 

\subsection{Compromised Operator and Threat Model}
\label{sub:operator}
 
A language-model operator \(\mathcal{O}\) has two roles in managing the shared
world state. At \(t=0\), it produces the mission allocation by assigning each
robot \(R_i\) a resource type \(\tau\), a pickup location
\(s_i\in\mathcal{P}\), and a drop-off location \(g_i\in\mathcal{D}\). Because
the allocation is not the object of study, one matching is fixed and reused
across all methods and iterations. At a later time \(t_H\), \(\mathcal{O}\)
receives and responds to incoming messages, creating a channel through which
malicious inputs can compromise its behavior 
\cite{perez2022ignore,greshake2023not,yang2024watch}.

Although manipulation can occur through multiple channels, we instantiate it
using an authority-framed prompt that asserts, without supporting physical
evidence, a finite set of obstacle cells
\(\mathcal{X}\subset\mathcal{F}\). We condition on successful manipulation so the operator accepts the claim. The planner treats \(\mathcal{X}\) as blocked when replanning against
the claim, but \(\mathcal{X}\) is never written to
\(G_{\mathrm{true}}\). Physical scans return \(G_{\mathrm{true}}(c)\), allowing
any robot that observes a claimed cell to provide evidence. 

\subsection{Claim Clusters and Impact}
\label{sub:impact}

\(\mathcal{X}\) is partitioned into maximal \(4\)-connected clusters \(\mathcal{Q}\), where each \(q\in\mathcal{Q}\) receives a binary impact value based on the remaining jobs at \(t_H\): \(I(q)=1\) if \(q\) intersects a remaining trajectory, and \(I(q)=0\) otherwise. Clusters with \(I(q)=1\) are prioritized for inspection. A cluster is marked \emph{seen} when all its cells have been observed.

\begin{figure*}[t]
    \centering
    \includegraphics[width=\linewidth]{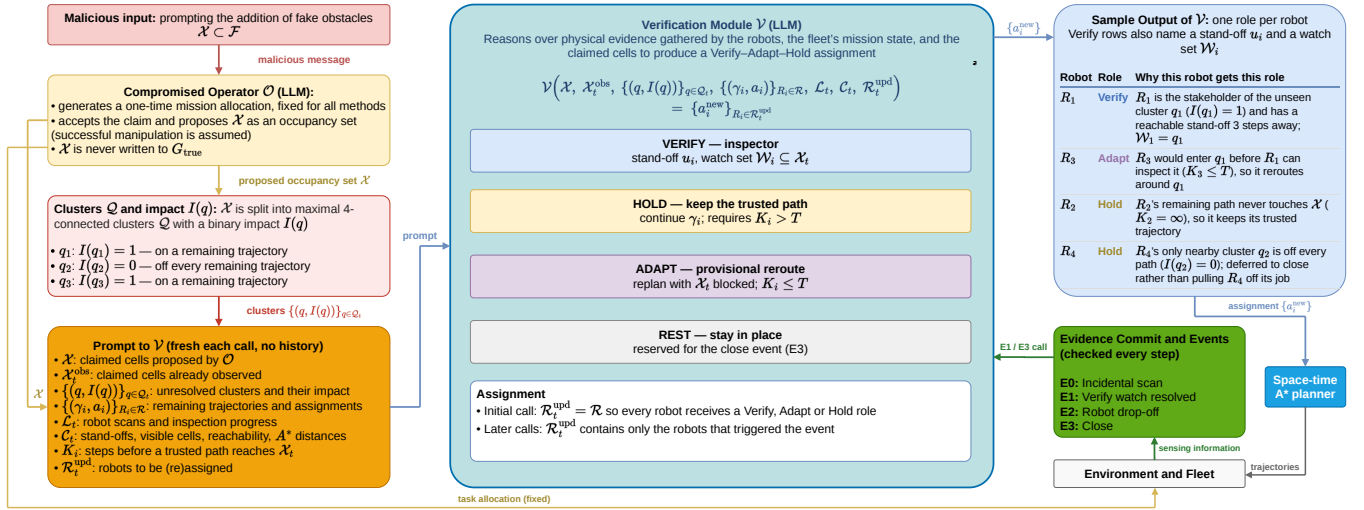}
\vspace{-15pt}
\caption{\textbf{Overview of the proposed framework:} A malicious input leads the
compromised operator $\mathcal{O}$ to accept fake obstacles $\mathcal{X}$ and
propose them as an occupancy set, which is never written to the trusted map.
The verification module $\mathcal{V}$, reasons over available physical evidence to produce
a Verify--Adapt--Hold assignment (with Rest reserved for the close step).
The initial call assigns every robot a verify, adapt or hold role and later calls reassign only the robots
that triggered an event. A space-time $A^{*}$ planner executes the assignment.}
\vspace{-15pt}
\label{fig:block_diag}
\end{figure*}

\section{Methodology}
\label{sec:method}

After the operator \(\mathcal{O}\) accepts a manipulated claim at \(t_H\),
our objective is to contain the physical consequences of its compromised
belief while preserving mission progress. We treat \(\mathcal{X}\) as a
provisional obstacle set and coordinate its verification with the remaining
transportation tasks. Robots receive one of three roles:
\emph{Verify} robots gather physical evidence to resolve claimed cells in
\(\mathcal{X}\); \emph{Adapt} robots provisionally reroute around unresolved
claimed cells when necessary to maintain mission progress; and \emph{Hold}
robots continue their trusted trajectories when immediate replanning is
unnecessary. These roles allow verification and mission execution to proceed
together while limiting unnecessary changes across the fleet.

Let \(\mathcal{V}\) denote the verification module that produces the Verify--Adapt--Hold assignments. The original external message is received only by \(\mathcal{O}\) and is
not passed directly to \(\mathcal{V}\). This separation limits direct
exposure to that message but does not guarantee correct verification
assignments. Figure~\ref{fig:block_diag} shows the overall framework.

\subsection{Unresolved Claims and Physical Evidence}
\label{sub:unresolved_claims}

The sets \(\mathcal{X}\) and \(\mathcal{Q}\) retain the original claimed
cells and their clusters. Let \(\mathcal{X}_t\subseteq\mathcal{X}\) contain
the claimed cells that remain unobserved after the scans at time \(t\).
Let \(\mathcal{S}_t\subseteq\mathcal{X}\) contain the claimed cells observed
by at least one robot at that time, according to the sensing range and
line-of-sight conditions in Sec.~\ref{sec:formulation}. Then
\vspace{-5pt}
\begin{equation}
\vspace{-5pt}
\mathcal{X}_t=
\begin{cases}
\mathcal{X}\setminus\mathcal{S}_t, & t=t_H,\\
\mathcal{X}_{t-1}\setminus\mathcal{S}_t, & t>t_H.
\end{cases}
\label{eq:unresolved_claim_update}
\end{equation}

Each scan supplies \(G_{\mathrm{true}}(c)\), and its evidence is committed
before any new role decision. In the evaluated setting,
\(\mathcal{X}\subseteq\mathcal{F}\), so observed claimed cells are
confirmed free and their provisional blocking status is removed. The claim
itself never changes \(G_{\mathrm{true}}\).

The unresolved clusters are $\mathcal{Q}_t=\left\{q\in\mathcal{Q}:q\cap\mathcal{X}_t\neq\varnothing\right\}$. For a cluster \(q\), only the cells in \(q\cap\mathcal{X}_t\) still require
inspection. Its original identity and impact \(I(q)\) are retained; it leaves
\(\mathcal{Q}_t\) only when all its cells have been observed. Thus,
\(\mathcal{X}_t\) shrinks as cells are scanned, while \(\mathcal{Q}_t\)
shrinks as clusters are fully resolved.

\subsection{Verify--Adapt--Hold Assignment}
\label{sub:vah_assignment}

Each call to \(\mathcal{V}\) uses a fresh prompt with no chat history.
Let
\(\mathcal{X}_t^{\mathrm{obs}}=\mathcal{X}\setminus\mathcal{X}_t\)
denote the claimed cells observed up to time \(t\).
Let \(\mathcal{L}_t\) denote the available robot scans and
\(\mathcal{C}_t\) denote the candidate stand-off information, including
the unresolved cells visible from each stand-off, its reachability,
and its geometric \(A^*\) distance for each robot. Reachability is
computed with \(\mathcal{X}_t\) and parked robots treated as walls.

For the current call, let
\(\mathcal{R}_t^{\mathrm{upd}}\subseteq\mathcal{R}\) denote the robots
requiring assignment updates, and let \(a_i^{\mathrm{new}}\) denote
the returned assignment for robot \(R_i\).
For a robot assigned \emph{Verify}, let
\(u_i\in\mathcal{F}\setminus\mathcal{X}_t\) denote its assigned
stand-off and \(\mathcal{W}_i\subseteq\mathcal{X}_t\) its watch set,
containing the unresolved claimed cells it is assigned to observe.
Its returned assignment is
\vspace{-5pt}
\begin{equation}
\vspace{-5pt}
a_i^{\mathrm{new}} = (s_i,g_i,\emph{Verify},u_i,\mathcal{W}_i).
\end{equation}
Assignments for the other roles retain \(s_i\) and \(g_i\) and specify
the selected role without a verification stand-off or watch set.
The module produces
\vspace{-5pt}
\begin{equation}
\vspace{-5pt}
\begin{aligned}
&\mathcal{V}\Bigl(
\mathcal{X},\,
\mathcal{X}_t^{\mathrm{obs}},\,
\{(q,I(q))\}_{q\in\mathcal{Q}_t},\\
&\qquad
\{(\gamma_i,a_i)\}_{R_i\in\mathcal{R}},\,
\mathcal{L}_t,\,
\mathcal{C}_t,\,
\mathcal{R}_t^{\mathrm{upd}}
\Bigr)\\
&\qquad =
\{a_i^{\mathrm{new}}\}_{R_i\in\mathcal{R}_t^{\mathrm{upd}}}.
\end{aligned}
\label{eq:vah_assignment}
\end{equation}
Initially, \(\mathcal{R}_t^{\mathrm{upd}}=\mathcal{R}\);
subsequent calls update only the robots associated with the
triggering event.

\noindent\textbf{Verify.} The assigned stand-off \(u_i\) must be reachable by robot \(R_i\), and
stand-offs assigned to different verifiers must be distinct. Every cell in \(\mathcal{W}_i\) must lie within sensing range of \(u_i\) and have an unobstructed line of sight from it. On the initial call, each unresolved cluster \(q\in\mathcal{Q}_t\) with \(I(q)=1\) receives a dedicated verifier satisfying \(q\cap\mathcal{X}_t \subseteq \mathcal{W}_i \subseteq \mathcal{X}_t\). Resolved clusters require no verifier, while clusters with \(I(q)=0\)
are deferred to the close event.

\noindent\textbf{Hold and Adapt.}
At a decision time \(t\), let \(k\geq 0\) index the remaining integer steps on a robot's trusted trajectory \(\gamma_i\). The number of steps before that trajectory first reaches an unresolved claimed cell is \(K_i = \min\left\{k\geq 0: p_i(t+k)\in\mathcal{X}_t\right\}\). Set \(K_i=\infty\) when the set is empty. Let \(T\) denote the longest travel time to a stand-off among the selected inspectors, with \(T=0\) when no inspector is required. For robots not assigned \emph{Verify}, the timing conditions are \(\emph{Hold}: K_i>T\) and \(\emph{Adapt}: K_i\leq T\).

\emph{Hold} preserves the trusted trajectory. If \(K_i\) is finite,
this choice also requires the relevant unresolved cells to be
assigned for inspection. \emph{Adapt} continues the remaining
mission along a replanned trajectory that treats \(\mathcal{X}_t\)
as blocked. Thus, holding means continuing the trusted plan,
while adaptation introduces a provisional detour.

\subsection{Event-Driven Role Updates}
\label{sub:vah_events}

After the initial assignment, the program checks four events at each step.
Let \(\mathcal{J}_t\subseteq\mathcal{R}\) denote the robots whose
pickup-and-delivery tasks remain incomplete at time \(t\).
Scan evidence and the resulting occupancy update are processed before
any event-driven model call. Only E1 and E3 call \(\mathcal{V}\).
Throughout a robot's current \emph{Verify} assignment, its watch set
\(\mathcal{W}_i\) is retained, while
\(\mathcal{W}_i\cap\mathcal{X}_t\) tracks the cells that still require
observation.

\noindent\textbf{E0: Incidental scan.}
Any robot, whether assigned \emph{Verify}, \emph{Hold}, or \emph{Adapt}, can provide new evidence through its scans. Using the observed-cell set \(\mathcal{S}_t\), new evidence about the claim is obtained when \(\mathcal{S}_t\cap\mathcal{X}_{t-1}\neq\varnothing\). This means that at least one previously unresolved claimed cell has been observed. If these observations complete no active verifier's watch set, the update is handled as E0. The remaining trajectories are replanned using the updated occupancy state, with every assignment \(a_i\) preserved and no call to \(\mathcal{V}\). If a verifier's watch set is completed, E1 applies to that verifier, regardless of which robot supplied the observations.

\noindent\textbf{E1: Verify watch resolved.}
For a robot currently assigned \emph{Verify}, its watch is resolved when \(\mathcal{W}_i\cap\mathcal{X}_t=\varnothing\). The observations may come from the inspector itself or another robot. The module is called for that inspector, with \(\mathcal{R}_t^{\mathrm{upd}}=\{R_i\}\). The returned assignment \(a_i^{\mathrm{new}}\) assigns it \emph{Adapt} to resume any unfinished part of its stored mission.

\noindent\textbf{E2: Robot drop-off.}
A holding or adapting robot completes its task while other tasks remain when \(R_i\in\mathcal{J}_{t-1}\setminus\mathcal{J}_t\) and \(\mathcal{J}_t\neq\varnothing\). It then becomes idle without a model call. Off-path clusters are left for the close event instead of immediately dispatching the completed robot.

\noindent\textbf{E3: Close.}
Let \(\mathcal{J}_t^{\mathrm{blocked}}\subseteq\mathcal{J}_t\) contain robots whose remaining pickup-and-delivery route is unreachable when \(\mathcal{X}_t\) is treated as walls. This concerns spatial reachability; temporary space-time reservation conflicts are handled by the planner. The close event applies when \(\mathcal{X}_t\neq\varnothing\) and \((\mathcal{J}_t=\varnothing \text{ or }\mathcal{J}_t^{\mathrm{blocked}}\neq\varnothing)\).

This allows inspection after delivery is complete or inspection
needed to release a mission blocked by an unresolved claim,
such as a sealed drop-off.
The module is called with the robots requiring closing assignments
in \(\mathcal{R}_t^{\mathrm{upd}}\).
Idle robots are assigned first; if none are available, the enclosed
robots inspect the region themselves. At the close step, at most
one robot is assigned to an unresolved cluster with \(I(q)=0\). \emph{Rest} is an additional role available only at the close event and is assigned to robots that are not sent for verification.

After each model call, the returned assignments are merged with
the retained assignments and validated before execution.
For each \(R_i\in\mathcal{R}_t^{\mathrm{upd}}\), the current
assignment \(a_i\) is replaced by \(a_i^{\mathrm{new}}\).
All other robots retain their existing assignments.

\subsection{Execution with the Shared Planner}
\label{sub:vah_execution}

We implement the space-time planner using \(A^*\). Following the initial assignment, \emph{Hold} trajectories
are reserved first. The planner then generates trajectories for
\emph{Verify} and \emph{Adapt} robots, treating unresolved claimed cells as
blocked for those roles. After an assignment change, only the affected robots receive new trajectories; all other trajectories are
fixed and reserved first. Following a scan that changes
\(\mathcal{X}_t\), all remaining trajectories are replanned against the
updated occupancy state. An \emph{Adapt} detour can therefore shorten as false claims are resolved.

If repeated reservations prevent progress, the planner retries the same
assignments with a different planning order. An assignment that remains
infeasible under the current obstacles and reservations is returned to the
module, which may propose another assignment for a bounded number of attempts.
The planner changes trajectory geometry and timing; it never changes the
assignment \(a_i\) itself.

\section{Experiments}
\label{sec:experiments}

\begin{figure}[h]
    \centering
    \includegraphics[width=0.45\textwidth]{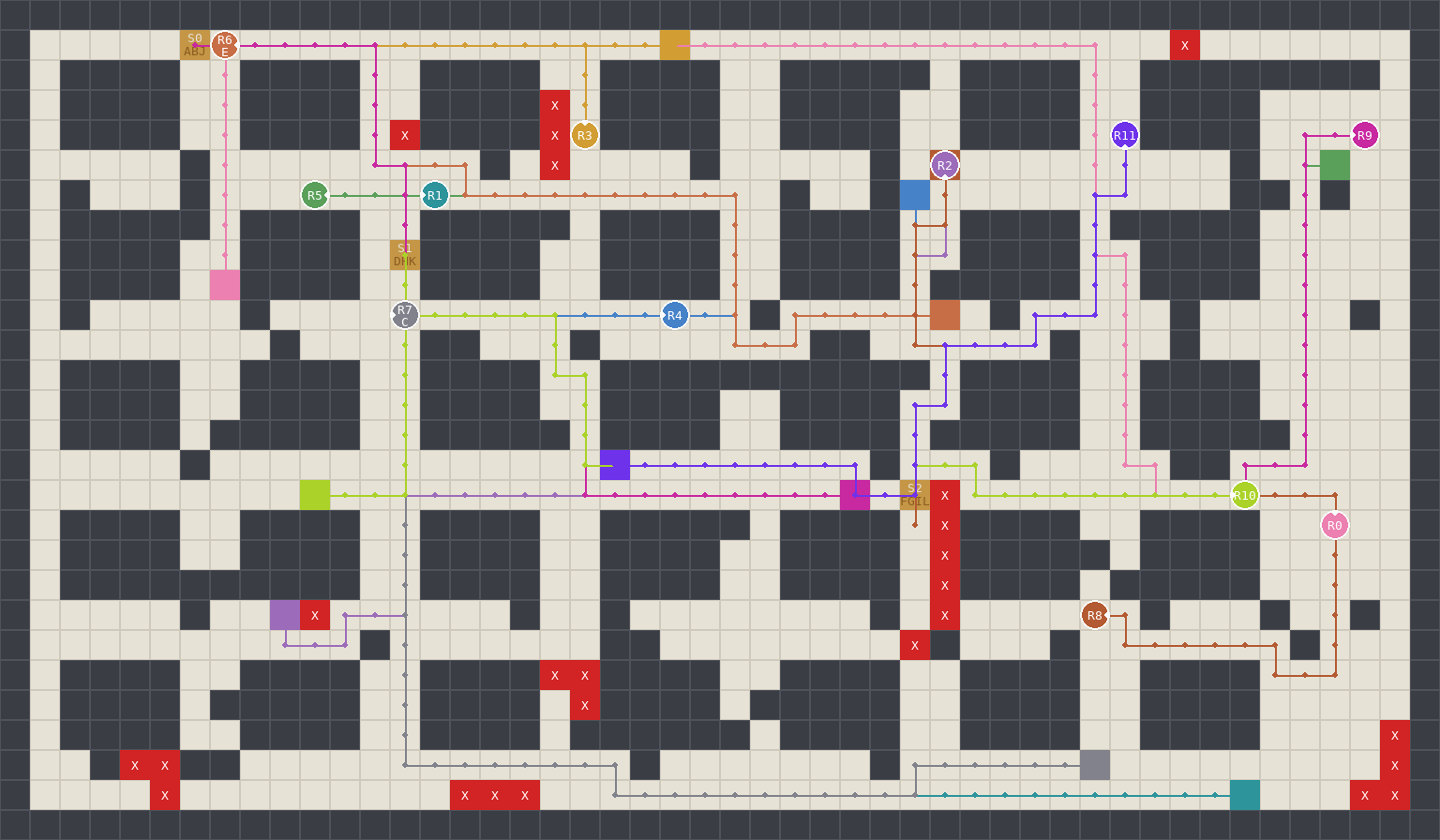}
    \caption{Warehouse simulation environment at time $t_H$ for $N=12$. \(S_0\), \(S_1\), and \(S_2\) denote the pickup stations, with four robots assigned to each station. Robots labeled with a letter have already collected their assigned resource. Red cells marked with \(\times\) represent the claimed obstacle set \(\mathcal{X}\). Each robot's drop-off location is indicated by the same color as the robot.}
    \label{fig:sim}
    \vspace{-15pt}
\end{figure}

Table~\ref{tab:experimental_setup} summarizes the experimental setup. The
mission allocation generated by \(\mathcal{O}\) remains fixed throughout each
experimental run. Figure \ref{fig:sim} shows a layout of the warehouse at time $t_H$ for $N=12$. 
\begin{table}[h]
\vspace{-5pt}
\centering
\caption{Experimental setup.}
\vspace{-5pt}
\label{tab:experimental_setup}
\footnotesize
\renewcommand{\arraystretch}{1.15}
\begin{tabular}{p{0.30\columnwidth} p{0.60\columnwidth}}
\hline
\textbf{Parameter} & \textbf{Configuration} \\
\hline
Team size
    & \(N\in\{3,6,9,12\}\) \\

Warehouse layout
    & Fixed across all experiments \\

Grid Size
    & $48 \text{ by } 28$ \\

Pickup locations
    & \(|\mathcal{P}|=3\) \\

Drop-off locations
    & \(|\mathcal{D}|=N\), with one unique drop-off per robot \\

Resources per pickup
    & \(N/3\) \\

Sensing
    & \(\rho=2\) \\

Operator model
    & \texttt{gpt-oss-120b} \\

Verification model
    & \texttt{gpt-oss-120b} \\
\hline
\end{tabular}
\vspace{-10pt}
\end{table}

\noindent\textbf{Claim Categories.}
We consider four claim categories spanning a controlled impact range:
\emph{Low}, \emph{High}, \emph{Mixed}, and \emph{Kill}. Each
\((N,\text{category})\) combination has \(M=15\) successful repetitions. Let
\(T_{\mathrm{nom}}\) denote the nominal makespan, defined as the number of
steps required to complete all jobs without a claim. The injection time
\(t_H\) is sampled uniformly from
\(\{1,\ldots,\lfloor T_{\mathrm{nom}}/5\rfloor\}\), corresponding to the first
fifth of the nominal mission. At \(t_H\), the false obstacle set
\(\mathcal{X}\) is generated and injected. This window allows a behavioral
cascade to develop and each method to respond. Within each repetition, all
methods use the same state and \(\mathcal{X}\).
\begin{enumerate}

\item \emph{Low.}
One isolated obstacle cell is placed on the trajectories of
\(k\sim\operatorname{Unif}\{1,\ldots,N\}\) robots selected without replacement.

\item \emph{High.}
One straight obstacle stick of length
\(\ell\sim\operatorname{Unif}\{2,\ldots,5\}\) is placed on every robot's
trajectory.

\item \emph{Mixed.}
The claim combines on-trajectory isolated obstacles and sticks with
off-trajectory obstacles. The numbers of isolated obstacles and sticks,
\((k_s,k_m)\), are sampled uniformly from the admissible integer pairs
satisfying \(0\leq k_s,k_m\leq\lceil N/2\rceil\) and
\(1\leq k_s+k_m\leq N\). It also includes
\(m_n\sim\operatorname{Unif}\{1,\ldots,\lceil N/2\rceil\}\) near-path
obstacles and
\(m_f\sim\operatorname{Unif}\{0,\ldots,\lceil N/2\rceil\}\) far,
wall-adjacent obstacles. Near-path obstacles may constrain passage, whereas
far obstacles affect no remaining trajectory. By combining consequential and
nonconsequential obstacles, this category tests whether a method prioritizes
verification of mission-threatening obstacles. Figure \ref{fig:sim} shows an instance of mixed obstacles injection.

\item \emph{Kill.}
A count \(k\sim\operatorname{Unif}\{1,\ldots,N\}\) is sampled, and \(k\)
robots with undelivered jobs are selected without replacement. Their drop-off
goals are then sealed by claimed obstacle cells.

\end{enumerate}

\subsection{Baselines}
\label{sec:baselines}

Every method is evaluated from the same fleet state at \(t_H\), with same $\mathcal{X}$, and remaining mission.

\noindent\textbf{Nominal.}
This is the no-claim reference. Each robot finishes its mission without incorporating \(\mathcal{X}\).

\noindent\textbf{Naive Replan.}
This method accepts all claimed cells as walls, setting
\(\mathcal{F}_i^{\mathrm{plan}}=\mathcal{F}\setminus\mathcal{X}\)
for every robot. Figure \ref{fig:sim} shows the rerouting around $\mathcal{X}$ under naive replan.

\noindent\textbf{Verification-First.}
This method selects inspectors through greedy set cover over the
candidate stand-offs. Let
\(\operatorname{Vis}(u)\subseteq\mathcal{X}\) denote the claimed
cells visible from stand-off \(u\) under the sensing model, and
let \(d_i(u)\) denote the geometric \(A^*\) distance from robot
\(R_i\) to \(u\).

Let \(\mathcal{U}\) contain the cells not yet covered by the
selected inspection assignments. Initially,
\(\mathcal{U}=\mathcal{X}_{t_H}\), so claimed cells already
observed from the robots' current positions require no dispatch.
At each greedy step, the method selects a feasible robot--stand-off
pair satisfying
\begin{equation}
(R_i,u_i)
\in
\operatorname*{arg\,max}_{(R_j,u)\ \mathrm{feasible}}
\left|\operatorname{Vis}(u)\cap\mathcal{U}\right|.
\label{eq:baseline_greedy_cover}
\end{equation}
Ties are broken by shorter \(A^*\) distance and then by lower
robot index. After selecting the pair, the uncovered set is updated:
\begin{equation}
\mathcal{U}
\leftarrow
\mathcal{U}\setminus\operatorname{Vis}(u_i).
\label{eq:baseline_cover_update}
\end{equation}
This update records planned inspection coverage.
The unresolved set \(\mathcal{X}_t\) changes only when cells
are physically observed. We consider two variants:

\noindent\textbf{Verify-Min.}
Let \(n_V\leq N\) denote the number of robots selected by the greedy cover. Only these \(n_V\) robots are dispatched for verification, while the remaining \(N-n_V\) robots halt until verification completes. Inspectors perform verification before resuming their remaining missions.

\noindent\textbf{Verify-All.}
All \(N\) robots are dispatched for verification. Each receives
a stand-off \(u_i\) satisfying
\(\operatorname{Vis}(u_i)\neq\varnothing\), even if its visible
cells are already covered by another inspection assignment.
The robots perform verification before resuming their missions.

\noindent\textbf{Rule-based VAH.}
This method uses the same inspector set as Verify-Min and assigns
the remaining robots using a fixed rule.
For each non-inspecting robot \(R_i\), let
\(\mathcal{B}_i\) contain the claimed cells on its remaining
trajectory:
\vspace{-5pt}
\begin{equation}
\vspace{-5pt}
\mathcal{B}_i
=
\left\{
c\in\mathcal{X}:
\exists\,t\geq t_H,\ (c,t)\in\gamma_i
\right\}.
\label{eq:baseline_path_intersection}
\end{equation}
Let \(f_i=1\) if all remaining pickup-and-delivery legs are
reachable in \(\mathcal{F}\setminus\mathcal{X}\), and let
\(f_i=0\) otherwise. These quantities are evaluated at \(t_H\).
The role component of \(a_i\) is selected as
\vspace{-5pt}
\begin{equation}
\vspace{-5pt}
\operatorname{role}(a_i)
=
\begin{cases}
\emph{Hold},
& \mathcal{B}_i=\varnothing,\\[1mm]
\emph{Adapt},
& \mathcal{B}_i\neq\varnothing,\ f_i=1,\\[1mm]
\emph{Stop},
& \mathcal{B}_i\neq\varnothing,\ f_i=0.
\end{cases}
\label{eq:baseline_rule_vah}
\end{equation}
A holding robot retains its trusted trajectory.
An adapting robot reroutes with \(\mathcal{X}\) treated as walls.
A stopped robot pauses its unfinished mission until verification
of the blocking region is performed. The initial role decision
is made once and is not changed throughout.

The verification-first baselines select inspectors by geometric
coverage. Rule-based VAH additionally considers trajectory
intersection and remaining-mission reachability.
These baselines do not use any way to defer unnecessary adaptation or revise role assignments.

\subsection{Metrics}
\label{sec:metrics}

Each method is evaluated from \(t_H\) onward against the \emph{Nominal} and
\emph{Naive Replan} references.

\noindent\textbf{Sum of Costs (SOC).}
For a method \(\pi\), let \(c(\pi_i)\) denote the number of cells traversed by
robot \(R_i\). The fleet-level cost is
\(\mathrm{SOC}=\sum_{i=0}^{N-1}c(\pi_i)\). Waiting does not contribute, so SOC
measures the total distance traveled by the fleet.

\noindent\textbf{Makespan.}
Makespan is the maximum number of time steps from \(t_H\) until the last robot
completes its remaining job. Unlike SOC, it includes time spent waiting.

\noindent\textbf{Coverage.}
Coverage is the fraction of claimed cells observed by at least one robot,
distinguishing methods that physically resolve the claim from those that merely
route around it.

SOC, makespan, and coverage are computed per run and averaged over the \(M\)
runs of each \((N,\text{category})\) condition.

\begin{figure*}[t]
    \centering
    \includegraphics[width=1\linewidth]{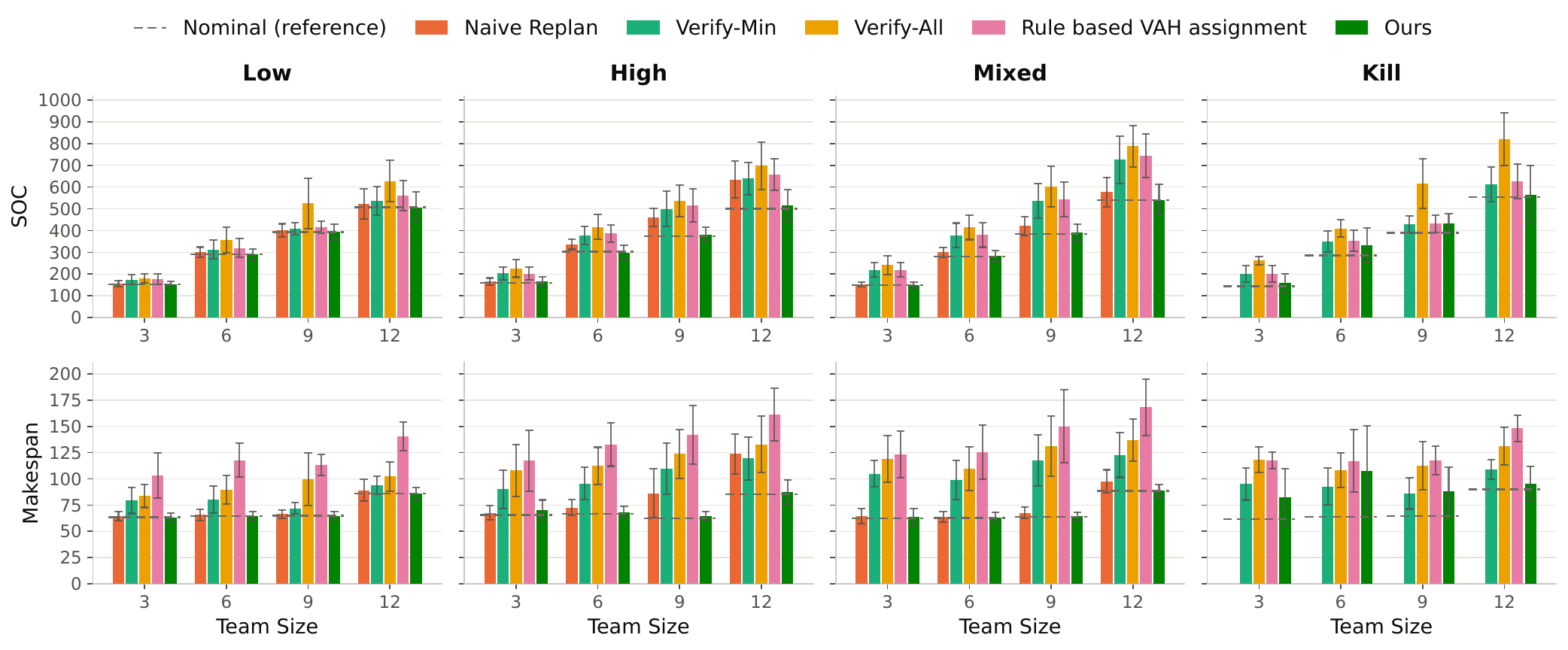}
    \vspace{-15pt}
    \caption{Sum of costs and Makespan variations across different scenarios and team sizes.}
    \vspace{-15pt}
    \label{fig:barcost}
\end{figure*}

\noindent\textbf{Cascade Containment Ratio.}
For a single run $m$, the additional distance traveled by method \(\pi\) relative
to Nominal is
\(D_m(\pi)=\sum_{i=0}^{N-1}\max\!\left(0,c(\pi_i)-c(\pi_i^{\mathrm{nom}})\right)\).
Because Naive Replan accepts the claim outright,
\(D_m(\pi^{\mathrm{naive}})\) represents the full cascade cost induced by the
claim. CCR is computed for each \((N,\text{category})\) condition by summing
the costs over its \(M\) runs before taking their ratio:
\begin{equation}
\mathrm{CCR}
=
1-
\frac{\sum_{m=1}^{M}D_m(\pi)}
{\sum_{m=1}^{M}D_m(\pi^{\mathrm{naive}})}.
\end{equation}

CCR measures the fraction of Naive Replan's aggregate excess travel avoided by a method. A value of \(1\) indicates no positive per-robot excess travel relative to Nominal, while \(0\) indicates the same aggregate excess travel as Naive Replan. CCR is bounded above by \(1\) but can be negative. It is undefined when the denominator is zero or a required Naive Replan cost is unavailable.

Two times are relevant: \(T_M\), when every remaining job has been delivered,
and \(T_V\), when every claimed cell has been scanned. Costs and CCR are
evaluated at \(T_M\), while coverage is evaluated at
\(\max(T_M,T_V)\), since verification may continue after mission completion. Thus, the main SOC and CCR results quantify travel up to mission completion and exclude any verification travel performed afterward.

\section{Results}
\label{sec:Results}

Figure~\ref{fig:barcost} shows that our method closely tracks the Nominal SOC
under Low, High, and Mixed impact, with its largest deviation under Mission
Kill. It achieves the lowest mean SOC in nearly every condition and remains
close to the lowest otherwise. All verification baselines exceed Nominal, with
Verify-All consistently incurring the greatest cost, and their gap from our
method generally grows with team size under High and Mixed impact. Our
method limits this cost by allowing non-inspecting robots to continue their
jobs and rerouting only those threatened by unresolved claims. Scanning a
claimed cell as free immediately shortens unnecessary detours, concentrating
verification cost among selected inspectors rather than propagating it across
the fleet. Mission Kill is the main exception because sealed drop-offs require
additional inspection travel before delivery.

\begin{figure}[t]
\centering
\includegraphics[width=0.9\columnwidth]{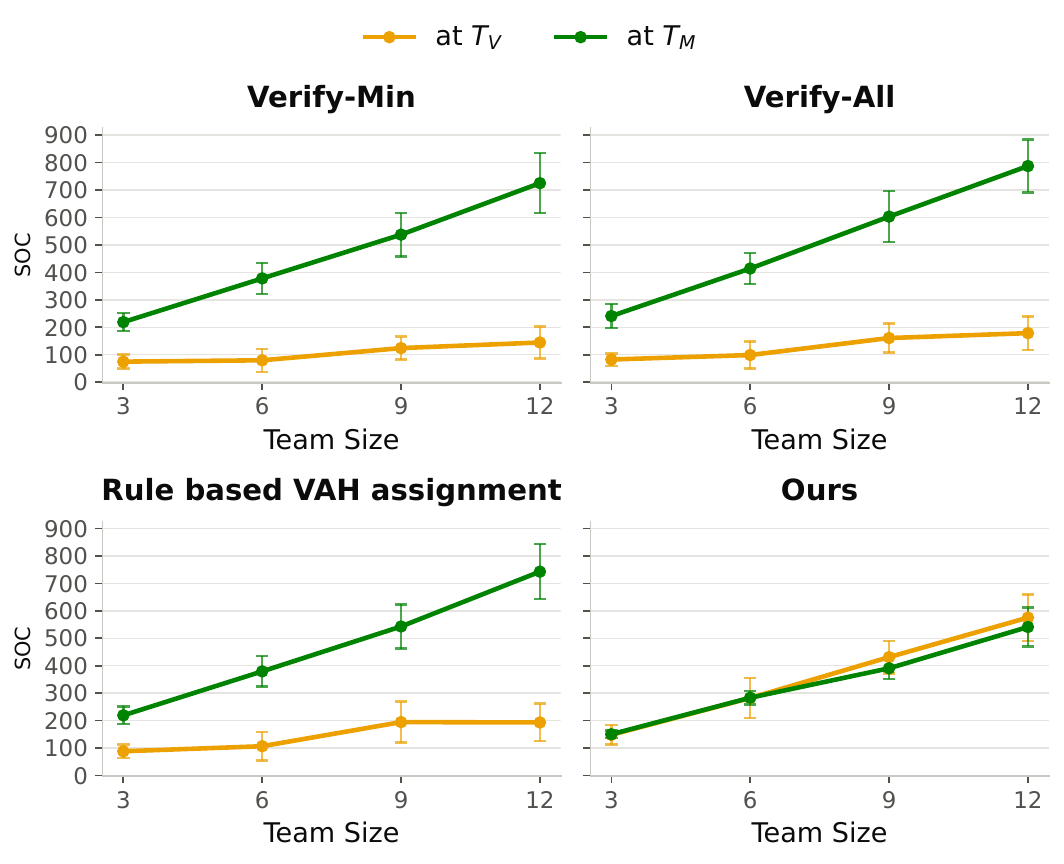}
\vspace{-5pt}
\caption{SOC at mission completion \(T_M\) and verification completion
\(T_V\) under Mixed-impact claims.}
\vspace{-15pt}
\label{fig:mixed_SOC_stamps}
\end{figure}

Figure~\ref{fig:mixed_SOC_stamps} compares SOC at mission completion, \(T_M\),
and verification completion, \(T_V\), under Mixed impact. For all verification
baselines, \(\mathrm{SOC}(T_M)>\mathrm{SOC}(T_V)\), and the gap grows with team
size. For our method, the values nearly coincide at smaller sizes, while
\(\mathrm{SOC}(T_M)<\mathrm{SOC}(T_V)\) becomes clearer at larger sizes.
This shows that our method can complete the mission before verification by resolving
consequential clusters first and deferring off-path cells until robots are no
longer required for delivery.

The makespan results follow a similar pattern. Our method remains close to
Nominal under Low, High, and Mixed impact and achieves the lowest or
near-lowest mean makespan outside Mission Kill. Verification-first increases
makespan by halting non-inspecting robots, while Rule-based VAH is slowest in
almost every condition despite having an SOC similar to Verify-Min. Its fixed
roles cause workers and inspectors to contend for corridors, serializing their
movements. Our selective, revisable assignments instead overlap verification
with mission execution. Under Mission Kill, Verify-Min inspects sealed drop-offs immediately, whereas
our method defers inspection until the close event determines that the
remaining jobs are infeasible. Ours is nonetheless fastest at \(N=3\), where
the enclosed robots inspect for themselves almost at once, and at \(N=12\),
where an idle robot is free the moment the close event fires; Verify-Min leads
at \(N=6\) and \(N=9\), where enough goals are sealed to delay completion but
too few robots are free to inspect them concurrently.

Table~\ref{tab:Coverage} reports coverage at \(N=12\), the largest fleet size. Every explicit-verification method, including ours, achieves full
coverage under all conditions except one. Our method under Mixed impact reaches
\(98\%\). This shortfall results from the method's inspection prioritization.
Off-path clusters are deferred until the close event. In this run, all
remaining unknown cells at the time of closure belonged to clusters that could
no longer affect the mission. The module therefore deemed them not worth a
dispatch and assigned the idle robots to \emph{Rest}. Thus, the method
prioritizes mission relevance over exhaustive coverage. Every mission-relevant
cluster was inspected throughout the study, and the off-path clusters were also
scanned in nearly every run. Naive Replan never achieves full coverage because
it observes cells only incidentally. It has no Mission Kill entry because
sealed goals are infeasible when claimed cells are treated as walls.

\begin{table}[t]
\caption{Coverage results for $N=12$ at $\max(T_M,T_V)$.}
\vspace{-5pt}
\setlength{\tabcolsep}{3pt}
\label{tab:Coverage}
\centering\scriptsize
\begin{tabular}{lccccc}
\toprule
Scenario & Naive & Verify-Min & Verify-All & Rule Based VAH & Ours \\
\midrule
Low   & $87\%$ & $100\%$ & $100\%$ & $100\%$ & $100\%$ \\
High  & $72\%$ & $100\%$ & $100\%$ & $100\%$ & $100\%$ \\
Mixed & $62\%$ & $100\%$ & $100\%$ & $100\%$ & $98\%$ \\
Kill  & \textemdash & $100\%$ & $100\%$ & $100\%$ & $100\%$ \\
\bottomrule
\end{tabular}
\vspace{-15pt}
\end{table}

Finally, Table~\ref{tab:ccr} shows that our method contains most of the cascade
in every reported condition except High impact at \(N=3\), where its CCR is
\(-0.65\). On all other occasions, CCR remains between \(0.60\) and \(1.00\),
whereas every verification baseline remains negative, incurring more
behavioral cascade than Naive Replan. The High-impact result at \(N=3\) reflects the cost of inspection in a fleet
too small to absorb it: every trajectory intersects the claim, so one robot
must be sent to a stand-off to view the full sticks while the other two wait or reroute, and with only
three robots that verification travel exceeds what Naive Replan spends
detouring around the sticks.

\begin{table}[H]
\vspace{-5pt}
\caption{Cascade containment ratio (CCR) results.}
\vspace{-5pt}
\label{tab:ccr}
\centering\scriptsize
\begin{tabular}{llcccc}
\toprule
Scenario & $N$ & Verify-Min & Verify-All & Rule Based VAH & Ours \\
\midrule
Low & 3 & $-3.89$ & $-5.14$ & $-4.06$ & $1.00$ \\
 & 6 & $-1.31$ & $-5.32$ & $-1.89$ & $1.00$ \\
 & 9 & $-1.36$ & $-17.18$ & $-2.22$ & $0.98$ \\
 & 12 & $-1.16$ & $-6.56$ & $-2.37$ & $0.96$ \\
\midrule
High & 3 & $-5.49$ & $-8.90$ & $-5.47$ & $-0.65$ \\
 & 6 & $-1.25$ & $-2.41$ & $-1.51$ & $0.81$ \\
 & 9 & $-0.50$ & $-0.91$ & $-0.67$ & $0.89$ \\
 & 12 & $-0.05$ & $-0.47$ & $-0.17$ & $0.83$ \\
\midrule
Mixed & 3 & $-16.60$ & $-22.10$ & $-16.63$ & $0.60$ \\
 & 6 & $-4.12$ & $-5.94$ & $-4.18$ & $0.83$ \\
 & 9 & $-3.22$ & $-5.01$ & $-3.35$ & $0.86$ \\
 & 12 & $-3.87$ & $-5.42$ & $-4.26$ & $0.95$ \\
\bottomrule
\end{tabular}
\vspace{-10pt}
\end{table}

\section{Conclusion}
\label{sec:conclusion}

This work shows that the physical consequences of a successfully manipulated 
LLM operator can be contained through decision-relevant team verification. The 
proposed \emph{Verify--Adapt--Hold} framework kept SOC and makespan near 
nominal in most conditions, achieved high cascade containment for larger 
fleets, and inspected every mission-relevant claim cluster while allowing 
unaffected robots to continue their tasks---demonstrating that post-compromise 
verification can prevent a false claim from causing fleet-wide disruption.

Future work will evaluate different manipulation channels, distribute role 
assignment across onboard LLMs in a decentralized fashion, validate the 
framework with larger fleets and real robots, and extend it to heterogeneous 
teams, limited communication, and partially valid claims.

\vspace{-2pt}

\section*{Acknowledgement}
Generative AI tools, including Cursor, ChatGPT, and Claude, assisted with
implementation, visualization, language refinement, and literature
organization. The authors developed and validated the research methodology,
experiments, analysis, and final manuscript.

\vspace{-1pt}
 

\bibliographystyle{ieeetr}
\bibliography{references}

\end{document}